\documentclass{article}
\usepackage{spconf}

\usepackage{cite}
\usepackage{amsmath,amssymb,amsfonts}
\usepackage{algorithmic}
\usepackage{graphicx}
\usepackage{textcomp}
\usepackage{xcolor}

\usepackage{array}
\usepackage{mathtools}
\usepackage{tcolorbox}
\usepackage{color}
\usepackage{theorem}
\usepackage{amssymb}
\usepackage{caption}
\usepackage{subcaption}
\usepackage{cite,hyperref}
\usepackage{cases}
\usepackage{url}
\usepackage{algorithmic}
\usepackage{algorithm}
\usepackage{tikz}
\usetikzlibrary{shapes,arrows}
\input{mysymbol.sty}
\input{my_beamer_defs.sty}

\usepackage{multirow}
\usepackage{hhline}

\newtcolorbox{myblockt}[1]{colback=urblue!5!white,
	colframe=urblue,fonttitle=\bfseries,
	title=#1}

\newtcolorbox{myblock}{colback=urblue!5!white,
	colframe=urblue,fonttitle=\bfseries}

\def\BibTeX{{\rm B\kern-.05em{\sc i\kern-.025em b}\kern-.08em
    T\kern-.1667em\lower.7ex\hbox{E}\kern-.125emX}}

\begin{document}
\ninept

\title{GraphMAD: Graph Mixup for Data Augmentation using Data-Driven Convex Clustering}
\name{Madeline Navarro and Santiago Segarra
\thanks{This work was supported by the NSF under award CCF-2008555.
Emails:  \href{mailto:nav@rice.edu}{nav@rice.edu}, \href{mailto:segarra@rice.edu}{segarra@rice.edu}}}
\address{Electrical and Computer Engineering, Rice University, USA}

\maketitle

\begin{abstract}%
We develop a novel \emph{data-driven nonlinear mixup mechanism for graph data augmentation} and present \emph{different mixup functions for sample pairs and their labels}.
Mixup is a data augmentation method to create new training data by linearly interpolating between pairs of data samples and their labels.
Mixup of graph data is challenging since the interpolation between graphs of potentially different sizes is an ill-posed operation. 
Hence, a promising approach for graph mixup is to first project the graphs onto a common latent feature space and then explore linear and nonlinear mixup strategies in this latent space.
In this context, we propose to (i) project graphs onto the latent space of continuous random graph models known as graphons, (ii) leverage convex clustering in this latent space to generate nonlinear data-driven mixup functions, and (iii) investigate the use of different mixup functions for labels and data samples.
We evaluate our graph data augmentation performance on benchmark datasets and demonstrate that nonlinear data-driven mixup functions can significantly improve graph classification.
\end{abstract}


\section{Introduction}

Rapid advancements in graph-based machine learning have led to the widespread use of graph neural networks (GNNs) in fields such as chemistry~\cite{gilmer2017neural}, wireless communications~\cite{chowdhury2021unfolding}, and social network analysis \cite{qiu2018deepinf}.
However, as with existing deep learning models, GNNs require large datasets to prevent overfitting.
Collecting more data is either costly or impossible, thus data augmentation arises as a cheap way to create more labeled samples from existing data.
New data are generated to preserve label-dependent characteristics while presenting unseen views of data to help the model learn discriminative features.

Mixup poses an efficient method for augmenting Euclidean data such as images and text \cite{lewy2022overview,guo2019augmenting}.
In essence, mixup creates new data by selecting points along the linear interpolants between pairs of existing labeled samples.
Formally, given two points $(\bbx_1,\bby_1)$ and $(\bbx_2,\bby_2)$ with data $\bbx_i$ and one-hot label vectors $\bby_i$ for $i=1,2$, mixup generates new samples as 
\begin{subequations}\label{e:orig_mix}
\begin{alignat}{2}
\bbx_{\mathrm{new}} = \lambda\bbx_1 + (1-\lambda)\bbx_2 \label{e:orig_featmix},\\
\bby_{\mathrm{new}} = \lambda\bby_1 + (1-\lambda)\bby_2 \label{e:orig_labelmix},
\end{alignat}
\end{subequations}
where the mixup parameter $\lambda\in[0,1]$ selects new points along the connecting line.
While mixup is theoretically and empirically effective at improving model generalization \cite{zhang2017mixup,guo2019mixup,zhang2021does}, it is not straightforward to translate conventional mixup to (non-Euclidean) graph data. 
Existing works \cite{guo2021intrusion,park2021graph,ding2022data,zhao2022graph} apply mixup in the graph domain by directly manipulating nodes, edges, or subgraphs; however, crucial structural information may be lost as classes can be sensitive to minor changes in graph topology.
An attractive alternative is mixing graph data in a latent space, which includes mixup of learned graph representations \cite{wang2021mixup,verma2021graphmix} or graph models \cite{han2022g}.

Traditional mixup of Euclidean data, including graph mixup in latent space, applies the same linear function for data mixup \eqref{e:orig_featmix} and label mixup \eqref{e:orig_labelmix} \cite{guo2020nonlinear}.
However, forcing linear mixing between classes may overlook crucial nonlinear behavior in the true distribution.
Additionally, given a pair of samples, the user must decide where along the linear interpolant to sample new data.
Previous works rely on empirically choosing the mixup parameter $\lambda$, but explicit investigation remains underexplored. 
Automatic data augmentation via learned optimal mixup parameters has attracted recent attention \cite{guo2020nonlinear,guo2019mixup,mai2021metamixup}, but this incurs great computational cost.

We present \emph{Graph Mixup for Augmenting Data (GraphMAD)} to augment graph data using data-driven mixup functions through convex clustering.
We also analyze the choice of different mixup functions for data [cf.~\eqref{e:orig_featmix}] and labels [cf.~\eqref{e:orig_labelmix}].
Our main contributions are as follows:
\bi[(i)] We perform \emph{nonlinear graph mixup in an interpretable continuous domain} given by $\emph{graphons}$, random graph models that characterize graph structure.
\i[(ii)] We present convex clustering to \emph{efficiently learn data-driven mixup functions}, where generated samples exploit relationships among all graphs as opposed to pairs of data.
\i[(iii)] We compare applying \emph{different mixup functions for data samples and their labels} and demonstrate examples of datasets for which this is beneficial.
\ei

We introduce preliminaries on graph mixup and convex clustering in Section \ref{S:prelims}.
Section \ref{S:method} describes the three main steps of our proposed GraphMAD process for augmenting graph data.
We demonstrate the superiority of our method in graph classification tasks on benchmark graph datasets in Section \ref{S:results}.
Finally, our paper concludes on a discussion of the future directions in Section \ref{S:concl}.

\begin{figure*}
\centering
	\begin{minipage}[c]{.9\textwidth}
		\includegraphics[width=\textwidth]{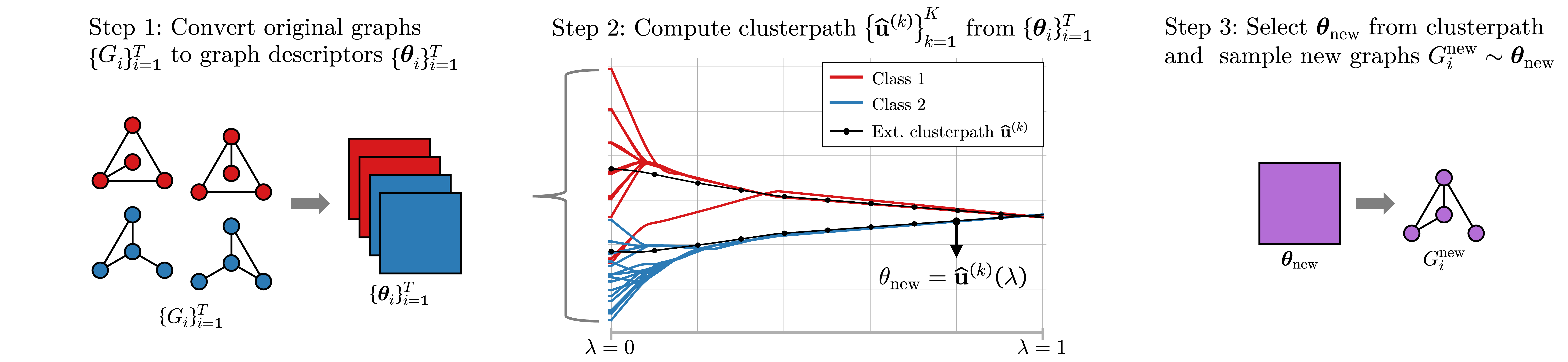}
	\end{minipage}
	\vspace{-2mm}
\caption{\small{Schematic of GraphMAD for graph data augmentation.
{\bf Step~1:} Convert each labeled graph $(G_i,\bby_i)$ to a labeled graph descriptor $(\bbtheta_i,\bby_i)$. {\bf Step~2:} Compute the clusterpath $\{\hat{\bbu}^{(k)}\}_{k=1}^K$ and soft labels $\{\hat{\bby}^{(k)}\}_{k=1}^K$ from the labeled descriptors $\{(\bbtheta_i,\bby_i)\}_{i=1}^T$. {\bf Step~3: } Sample $\bbtheta_{\mathrm{new}}$ from a point in the clusterpath $\hat{\bbu}^{(k)}(\lambda)$ at $\lambda\in[0,1]$ and $k=1,\dots,K$ and generate new graphs $G_i^{\mathrm{new}}$ with label $\bby_{\mathrm{new}} = \hat{\bby}^{(k)}(\lambda)$.
}}
	\vspace{-4mm}
\label{fig:graphmad}
\end{figure*}

\section{Preliminaries and Related Work}\label{S:prelims}


In this section, we introduce three key topics related to our proposed GraphMAD method.
We review mixup for graph data, graphons as continuous graph descriptors, and convex clustering.

\subsection{Graph Mixup}\label{Ss:graphmixup}

Since its introduction \cite{zhang2017mixup}, mixup has been popularly applied to image and text data~\cite{lewy2022overview,guo2019augmenting}.
The application of mixup to graphs is a new but quickly developing area of research due to the ubiquity of graph data.
Early research on graph mixup operates in the graph domain, directly modifying the topology of the graphs \cite{guo2021intrusion,park2021graph}.
However, the cascading nature of graph connectivity means that crucial class-dependent structural information can be lost by changing a single edge or node.
Thus, mixup for graphs in a latent graph representation space has gained popularity \cite{wang2021mixup,verma2021graphmix,han2022g}.

Authors in \cite{han2022g} present the closest work to our own. 
They estimate a set of $K$ nonparametric random graph models known as \emph{graphons} to collect discriminative characteristics for each of the $K$ classes.
Pairs of class graphons are then linearly interpolated to obtain new random graph models to create any number of new graph samples.
However, because a single graphon represents all graphs in one class, the subsequent mixup cannot consider the spread of graphs within each class.
Moreover, they consider only linear interpolation.

\subsection{Continuous Graph Descriptors}\label{Ss:graphdesc}

To perform mixup of graphs in a continuous latent space as in \cite{wang2021mixup,verma2021graphmix,han2022g}, we require an Euclidean graph descriptor.
In this work, similar to \cite{han2022g}, we adopt the \emph{graphon}, a bounded symmetric function $\ccalW:[0,1]^2\rightarrow[0,1]$ that can be interpreted as a random graph model associated with a family of graphs with similar structural characteristics \cite{lovasz2012large,diaconis2007graph,avella2020centrality}.
To generate an undirected, unweighted graph $G=(\ccalV,\ccalE)$ from a graphon $\ccalW$, we require the following two steps
\begin{subequations}\label{equ_graphon_samp}
\begin{alignat}{2}
&\zeta_{i} \sim \text{Uniform}([0,1]) &&\forall~i\in\ccalV,
\label{equ_graphon_samp1}\\
&\bbA_{ij} = \bbA_{ji} \sim \text{Bernoulli}\left( \ccalW(\zeta_i,\zeta_j) \right) &\qquad& \forall~(i,j)\in\ccalV\times\ccalV,
\label{equ_graphon_samp2}
\end{alignat}
\end{subequations}
where $\bbA\in\{0,1\}^{|\ccalV|\times|\ccalV|}$ is the adjacency matrix with $\bbA_{ij}\neq 0$ if and only if $(i,j)\in\ccalE$, and the latent variables $\zeta_i\in[0,1]$ are independently drawn for each node $i$.
Intuitively, each node $i$ is assigned a value $\zeta_i$ and the edge $(i,j)$ is drawn between nodes $i$ and $j$ with probability $\ccalW(\zeta_i,\zeta_j)$ for all $i\neq j$.

We convert every graph to a descriptor via graphon estimation, a well-studied task.
A graphon can be inferred as a continuous function \cite{sischka2022based,chan2014consistent} or as a coarser piecewise-constant stochastic block model (SBM) \cite{chan2013estimation}.

\begin{table*}[t]
\scriptsize
\centering
\caption{\small{Graph classification accuracy on molecule and bioinformatics datasets. 
Columns $g_{\mathrm{feat}}$ and $g_{\mathrm{label}}$ are the data mixup for \eqref{e:mixup_feat} and the label mixup for \eqref{e:mixup_label}, respectively.
The top performing methods are {\bf bolded}.
}}
\vspace{-2mm}
\begin{tabular}{ m{6em} m{6.5em} m{6em} m{6em} m{6em} m{6em} m{6.5em} m{6em} }
\hline
\multicolumn{2}{c}{\bf Method} & \multicolumn{1}{c}{\bf DD} & \multicolumn{1}{c}{\bf PROTEINS} & \multicolumn{1}{c}{\bf ENZYMES} & \multicolumn{1}{c}{\bf AIDS} & \multicolumn{1}{c}{\bf MUTAG} & \multicolumn{1}{c}{\bf NCI109} \\
\hline
\multicolumn{2}{c}{Classes} & \multicolumn{1}{c}{2} & \multicolumn{1}{c}{2} & \multicolumn{1}{c}{6} & \multicolumn{1}{c}{2} & \multicolumn{1}{c}{2} & \multicolumn{1}{c}{2} \\
\multicolumn{2}{c}{Graphs} & \multicolumn{1}{c}{1178} & \multicolumn{1}{c}{1113} & \multicolumn{1}{c}{600} & \multicolumn{1}{c}{2000} & \multicolumn{1}{c}{188} & \multicolumn{1}{c}{4127} \\
\hline
$g_{\mathrm{feat}}$ & $g_{\mathrm{label}}$ & & & & & & \\
\hhline{|-|-|}
\textemdash & \textemdash & $68.77\pm2.35$ & $\mathbf{69.51\pm1.20}$ & $26.43\pm2.55$ & $96.18\pm2.57$ & $84.59\pm5.53$ & $68.23\pm2.13$ \\[.07cm]
\multirow{4}{*}{Linear \eqref{e:gmixup}} & Linear \eqref{e:orig_labelmix}~\cite{han2022g} & $67.01\pm1.72$ & $65.15\pm2.53$ & $24.88\pm3.38$ & $96.82\pm1.39$ & $85.71\pm7.15$ & $68.16\pm2.72$ \\
& Sigmoid \eqref{e:mixup_label_sig} & $64.89\pm1.49$ & $68.42\pm3.94$ & $24.76\pm4.10$ & $96.07\pm1.42$ & $85.71\pm4.63$ & $65.96\pm2.34$ \\
& Logit \eqref{e:mixup_label_log} & $66.22\pm3.82$ & $69.25\pm2.94$ & $25.95\pm5.48$ & $96.07\pm1.27$ & $80.08\pm5.60$ & $66.81\pm4.07$ \\
& Clusterpath \eqref{e:gmad_label} & $68.22\pm3.71$ & $69.38\pm2.04$ & $24.64\pm2.39$ & $95.86\pm1.88$ & $\mathbf{87.22\pm4.96}$ & $65.01\pm3.07$ \\[.07cm]
\multirow{4}{*}{Clusterpath \eqref{e:gmad_feat}} & Linear \eqref{e:orig_labelmix} & $67.11\pm1.56$ & $67.51\pm2.62$ & $\mathbf{26.67\pm6.49}$ & $\mathbf{97.15\pm1.00}$ & $\mathbf{87.24\pm4.21}$ & $\mathbf{68.61\pm1.41}$ \\
& Sigmoid \eqref{e:mixup_label_sig} & $68.23\pm3.61$ & $64.60\pm5.07$ & $\mathbf{32.62\pm6.35}$ & $97.07\pm1.35$ & $85.20\pm3.53$ & $67.50\pm2.06$ \\
& Logit \eqref{e:mixup_label_log} & $\mathbf{70.07\pm2.51}$ & $67.26\pm2.84$ & $25.71\pm4.26$ & $95.87\pm1.47$ & $80.10\pm14.77$ & $65.33\pm3.35$ \\
& Clusterpath \eqref{e:gmad_label} & $\mathbf{70.44\pm3.79}$ & $\mathbf{71.18\pm3.98}$ & $24.52\pm3.30$ & $\mathbf{97.22\pm0.54}$ & $85.71\pm5.40$ & $\mathbf{68.54\pm3.16}$ \\
\hline
\end{tabular}\label{t:molbio}
\vspace{-4mm}
\end{table*}

\subsection{Convex Clustering}\label{Ss:cvxclust}

We apply convex clustering to characterize the spatial relationship of the graph data in graphon space.
Convex clustering formulates the clustering problem as a continuous optimization problem, first introduced by \cite{pelckmans2005convex,hocking2011clusterpath,lindsten2011clustering} and inspired by sum-of-norms regularization \cite{ohlsson2010segmentation}.
Given a set of $T$ data samples $\bbx = \{\bbx_i\}_{i=1}^T$ in $\mathbb{R}^p$, traditional convex clustering seeks to solve the regularized problem
\begin{alignat}{3}
\hat{\bbu}(\gamma) = \argmin_{ \bbu } \sum_{i=1}^T d_{\mathrm{fid}}(\bbu_i,\bbx_i) + \gamma\sum_{i<j} w_{ij}d_{\mathrm{fus}}(\bbu_i,\bbu_j), \label{e:cvxclust}
\end{alignat}
where $\gamma\geq0$ is the tunable fusion parameter, $w_{ij}\geq0$ is the weight specifying the fusion strength between $\bbu_i$ and $\bbu_j$, and the functions $d_{\mathrm{fid}}$ and $d_{\mathrm{fus}}$ quantify data fidelity and sample fusion, respectively.
We let the \emph{clusterpath} be the solution path $\hat{\bbu} = \{\hat{\bbu}_i\}_{i=1}^T$ providing the cluster centroids $\hat{\bbu}(\gamma) = \{\hat{\bbu}_i(\gamma)\}_{i=1}^T$ for any $\gamma\geq0$.
If $\hat{\bbu}_i(\gamma) = \hat{\bbu}_j(\gamma)$ for $i\neq j$, then we say that samples $\bbx_i$ and $\bbx_j$ belong to the same cluster at $\gamma$, where $\hat{K}(\gamma)$ is the total number of clusters for the fusion level $\gamma$.
We show an example clusterpath in Step 2 of Fig. \ref{fig:graphmad}.

Under mild conditions, convex clustering enjoys theoretical guarantees for agglomerative fusions and cluster recovery due to the convexity of \eqref{e:cvxclust} \cite{tan2015statistical,zhu2014convex,chi2019recovering}.
However, convex clustering in non-Euclidean domains remains underexplored \cite{choi2019convex,navarro2021network}.
Indeed, applying \eqref{e:cvxclust} for graph data suffers from the same difficulties as mixup of graphs, and imposing valid data formats may require nonconvex penalties, thus obviating the value of convex clustering.

\section{Methodology}\label{S:method}

We let $G = (\ccalV,\ccalE)$ denote an undirected, unweighted graph with set of nodes $\ccalV$ and set of edges $\ccalE = \ccalV\times \ccalV$. 
For a labeled graph $G$ belonging to one of $K$ classes, its label is represented by the one-hot vector $\bby\in\{0,1\}^K$.
When $G$ belongs to the $k$-th class, $\bby$ contains zeros everywhere except the $k$-th entry, which has the value 1.
We denote the space of all graphs by $\ccalG$.

\subsection{GraphMAD}\label{Ss:graphmad}

Given a set of $T$ samples $\{(G_i,\bby_i)\}_{i=1}^T$ of graph labeled data used to train a classifier $f: \ccalG \rightarrow\{0,1\}^K$, our goal is to use GraphMAD to generate new data $\{(G_i^{\mathrm{new}},\bby_i^{\mathrm{new}})\}_{i=1}^{T'}$ so that a classifier ${f_{\mathrm{new}}}$ trained on the augmented dataset $\{(G_i,\bby_i)\}_{i=1}^T\cup\{(G_i^{\mathrm{new}},\bby_i^{\mathrm{new}})\}_{i=1}^{T'}$ outperforms $f$ when classifying unseen graphs.
In particular, we focus on a GNN graph classifier \cite{kipf2016semi,gilmer2017neural,xu2018powerful}, a highly effective model that requires a large training dataset.

We perform GraphMAD in graph descriptor space to obviate the difficulties of graph data for convex clustering and mixup described in Section \ref{S:prelims}.
More specifically, we convert each graph $G_i$ to an SBM graphon estimate \cite{chan2014consistent} as introduced in Section \ref{Ss:graphdesc}.
Given a set of graph descriptors denoted by $\bbtheta = \{\bbtheta_i\}_{i=1}^T$ and their respective one-hot labels $\bby = \{\bby_i\}_{i=1}^T$, we formulate the general mixup process of graph data as 
\begin{subequations}\label{e:mixup}
\begin{alignat}{2}
&\bbtheta_{\mathrm{new}} = g_{\mathrm{feat}}(\bbtheta;\lambda), \label{e:mixup_feat}\\
&\bby_{\mathrm{new}} = g_{\mathrm{label}}(\bbtheta,\bby;\lambda) \label{e:mixup_label},
\end{alignat}
\end{subequations}
where $\lambda\in[0,1]$ is the mixup parameter and $g_{\mathrm{feat}}$ and $g_{\mathrm{label}}$ are the mixup functions for graph descriptors and labels, respectively.
For traditional mixup, we have that $g_{\mathrm{feat}}(\bbx;\lambda) = \lambda\bbx_i + (1-\lambda)\bbx_j$ and $g_{\mathrm{label}}(\bbx,\bby;\lambda) = \lambda\bby_i + (1-\lambda)\bby_j$ for $i,j=1,\dots,T$, where $i\neq j$. 

A schematic of our graph mixup method is presented in Fig. \ref{fig:graphmad}, which can be described in three steps: {\bf Step~1:} We convert each graph $G_i$ to a descriptor $\bbtheta_i=\bbtheta(G_i)$; {\bf Step~2:} Given the descriptors $\bbtheta$, we compute the clusterpath $\hat{\bbu}$ over $\lambda\in[0,1]$ through convex clustering; and {\bf Step~3:} We choose $\lambda\in[0,1]$ to select points in the clusterpath $\bbtheta_{\mathrm{new}} = \hat{\bbu}(\lambda)$. 
The samples $\bbtheta_{\mathrm{new}}$ in graph descriptor space are converted to the graph domain by sampling $G_i^{\mathrm{new}}\sim \bbtheta_{\mathrm{new}}$ from the graphon with label $\bby_{\mathrm{new}}$.
The remainder of this section elaborates on each step of GraphMAD.

\medskip
\noindent{\bf Step 1: Graph Descriptors.}
Convex clustering in Section \ref{Ss:cvxclust} requires Euclidean data formats for the fidelity and fusion distances in \eqref{e:cvxclust}.
We convert the non-Euclidean graph data $\{G_i\}_{i=1}^T$ to descriptors from Section \ref{Ss:graphdesc}, denoted as $\bbtheta_i=\bbtheta(G_i)$ for samples $i=1,\dots,T$.
Similarly to \cite{han2022g}, our graph descriptors are SBM graphon approximations, where $\bbW_i\in[0,1]^{D\times D}$ is obtained for each graph $G_i$ by sorting and smoothing (SAS)  \cite{chan2014consistent} with $D$ denoting the fineness of the graphon estimate.
The graph descriptors are then set as $\bbtheta_i = \text{vec}(\bbW_i) \in[0,1]^{D^2}$, where $\text{vec}(\bbY)$ denotes the vectorization of the matrix $\bbY$.
In this paper, we restrict our analysis to graphons, but GraphMAD accepts any descriptors that permit computation of \eqref{e:cvxclust} given choices of $d_{\mathrm{fid}}$ and $d_{\mathrm{fus}}$.

\medskip
\noindent{\bf Step 2: Data-Driven Mixup Function via Clusterpath.}
We approach graph mixup with the goal of exploiting the positions of data samples to inform where and how to interpolate between classes.
We characterize the spatial relationships of graphs in descriptor space through the clusterpath, which we obtain by adapting \eqref{e:cvxclust} as
\begin{alignat}{3}
\hat{\bbu}(\lambda) =  \argmin_{\bbu} ~ \sum_{i=1}^T \|\bbu_i-\bbtheta_i\|_2^2 + \frac{\lambda}{1-\lambda}\sum_{i<j} w_{ij}\|\bbu_i-\bbu_j\|_1, \label{e:graphmad}
\end{alignat}
where we specified the fidelity $d_{\mathrm{fid}}$ and fusion $d_{\mathrm{fus}}$ distances as the squared $\ell_2$-norm and $\ell_1$-norm, respectively, and we have $\lambda\in[0,1]$ as in traditional mixup.
Section \ref{Ss:cvxclust} describes the clusterpath $\hat{\bbu}$ as providing graphons $\hat{\bbu}(\lambda)$ for $\lambda\in[0,1]$, where $\hat{\bbu}_i(0) = \bbtheta_i$ returns the original data and $\hat{\bbu}_i(1) = \frac{1}{T}\sum_{j=1}^T \bbtheta_j$ results in total fusion of the dataset for all $i=1,\dots,T$.
Furthermore, we let the fusion weights $w_{ij} = 1$ when $G_i$ and $G_j$ belong to the same class, and $w_{ij} = \epsilon<1$, otherwise.
The labels inform the fusion weights to encourage tree-like clustering \cite{chi2019recovering} while maintaining the data-dependent clusterpath shape.
Note that the solutions to \eqref{e:graphmad} lie in the convex hull of $\bbtheta$, resulting in valid symmetric and bounded graphons described in Section \ref{Ss:graphdesc} and Step 1 of Section \ref{Ss:graphmad}.


The clusterpath $\hat{\bbu}=\{\hat{\bbu}_i\}_{i=1}^T$ obtained via \eqref{e:graphmad} results in $T$ graphons that vary over $\lambda\in[0,1]$.
We compute a clusterpath of $K$ branches by combining subsets of the $T$ paths based on their cluster assignments from \eqref{e:graphmad}.
Given a point $\lambda^*$ where the number of clusters $\hat{K}(\lambda^*) = K$ as in Section \ref{Ss:cvxclust}, let the cluster assignments of $\hat{\bbu}(\lambda^*)$ be represented by index sets $\ccalI^{(k)}\subseteq\{1,\dots,T\}$ for $k=1,\dots,K$, where $\ccalI^{(1)}\cup \cdots \cup \ccalI^{(K)} = \{1,\dots,T\}$.
We collapse $\hat{\bbu}$ into $K$ branches by averaging $K$ graphon subsets based on $\{\ccalI^{(k)}\}_{k=1}^K$ as
\begin{equation}\label{e:graphmad_cp}
\hat{\bbu}^{(k)}(\lambda) = \frac{1}{|\ccalI^{(k)}|} \sum_{i\in\ccalI^{(k)}} \hat{\bbu}_i(\lambda) \quad \forall k=1,\dots,K,~\lambda\in[0,1].
\end{equation}
The \emph{extended clusterpath} $\{\hat{\bbu}^{(k)}\}_{k=1}^K$ is superimposed on the original clusterpath in Step 2 of Fig.~\ref{fig:graphmad}.

\medskip
\noindent{\bf Step 3: Sampling.}
In this step, we describe the mixup processes in \eqref{e:mixup} using the extended clusterpath in \eqref{e:graphmad_cp} for sampling new data $g_{\mathrm{feat}}$ and new labels $g_{\mathrm{label}}$.

\medskip
\noindent{\bf New data $\bbtheta_{\mathrm{new}}$.}
To obtain $\bbtheta_{\mathrm{new}} = g_{\mathrm{feat}}(\bbtheta;\lambda)$ from the extended clusterpath $\{\hat{\bbu}^{(k)}\}_{k=1}^K$ in \eqref{e:graphmad_cp}, we first choose a branch $k\in\{1,\dots,K\}$ and mixup parameter $\lambda\in[0,1]$, and we obtain the graphon centroid as 
$\bbtheta_{new} = \hat{\bbu}^{(k)}(\lambda)$.
A set of $T'$ new graphs $\{G_i^{\mathrm{new}}\}_{i=1}^{T'}$ is created from the graphon $\bbtheta_{new}$ as
\begin{equation}\label{e:gmad_feat}
G_i^{\mathrm{new}}\sim \bbtheta_{\mathrm{new}} = \hat{\bbu}^{(k)}(\lambda) 
\end{equation}
where $G_i^{\mathrm{new}}\sim \bbtheta_{\mathrm{new}}$ refers to the process of sampling a graph from a graphon as described in \eqref{equ_graphon_samp}, so we can generate any number of new graphs for every $k$ and $\lambda$ that we consider.

\medskip
\noindent{\bf New labels ${\bby}_{\mathrm{new}}$.}
We can obtain labels ${\bby}_{\mathrm{new}} = g_{\mathrm{label}}(\bbtheta,\bby;\lambda)$ using the extended clusterpath $\{\hat{\bbu}^{(k)}\}_{k=1}^K$ as with $\bbtheta_{\mathrm{new}}$.
To do this, we compute soft labels ${\bby}_{\mathrm{new}}\in[0,1]^K$ using class proportions of samples $\bbtheta$ in each branch of the clusterpath $\hat{\bbu}^{(k)}$, as described below.

For each branch, we define a label clusterpath function $\hat{\bby}^{(k)}(\lambda) = g_{\mathrm{label}}(\bbtheta,\bby;\lambda)$ that yields the soft label ${\bby}_{\mathrm{new}} = \hat{\bby}^{(k)}(\lambda) \in[0,1]^K$ corresponding to the graphon $\hat{\bbu}^{(k)}(\lambda)$.
For the $k$-th branch, we let the $l$-th entry of $\hat{\bby}^{(k)}(0)\in[0,1]^K$ be the proportion of class $l$ assigned to branch $k$, and we let $\hat{\bby}^{(k)}(1) = \frac{1}{K}{\bf 1}$, where ${\bf 1}$ is the all-ones vector of length $K$, denoting total dataset fusion for equally-sized classes.
We then compute $\hat{\bby}^{(k)}(\lambda)$ for $\lambda\in[0,1]$ as 
\begin{equation}\label{e:gmad_label}
\hat{\bby}^{(k)}(\lambda) = g_{\mathrm{cp}}^{(k)}(\lambda)\hat{\bby}^{(k)}(0)
+ \left(1-g_{\mathrm{cp}}^{(k)}(\lambda)\right)\hat{\bby}^{(k)}(1),
\end{equation}
where $g_{\mathrm{cp}}^{(k)}$ represents the rate of change of the $k$-th branch of the clusterpath $\hat{\bbu}^{(k)}$ with respect to the mixup parameter $\lambda$.
More specifically, we have $g_{\mathrm{cp}}^{(k)}:[0,1]\rightarrow[0,1]$ as
\begin{equation}\label{e:cp_roc}
    g_{\mathrm{cp}}^{(k)}(\lambda) = \frac{\|\hat{\bbu}^{(k)}(0)+\int_0^{\lambda}\nabla_{\lambda}\hat{\bbu}^{(k)}(\tau)d\tau\|_2^2-\|\hat{\bbu}^{(k)}(0)\|_2^2}{\|\hat{\bbu}^{(k)}(1)\|_2^2-\|\hat{\bbu}^{(k)}(0)\|_2^2},
\end{equation}
for $\lambda\in[0,1]$ and $k=1,\dots, K$, where $g_{\mathrm{cp}}^{(k)}(\lambda) = \lambda$ for $\lambda\in\{0,1\}$.

Similarly to $\bbtheta_{\mathrm{new}} = \hat{\bbu}^{(k)}(\lambda)$, new graph labels ${\bby}_{\mathrm{new}}=\hat{\bby}^{(k)}(\lambda)$ for $\lambda\in[0,1]$ require selecting a branch $k\in\{1,\dots,K\}$. 
When the clusterpath is used for both $g_{\mathrm{feat}}$ and $g_{\mathrm{label}}$, that is, $\bbtheta_{\mathrm{new}} = \hat{\bbu}^{(k)}(\lambda)$ in \eqref{e:gmad_feat} and ${\bby}_{\mathrm{new}} = \hat{\bby}^{(k)}(\lambda)$ in \eqref{e:gmad_label}, we use the same branch $k$ of the extended clusterpath $\hat{\bbu}^{(k)}$ for \eqref{e:gmad_feat} and \eqref{e:cp_roc}.

\medskip
Under Steps 1 to 3, we obtain data-driven mixup functions for \eqref{e:mixup} using descriptors $\bbtheta$, where new samples $\bbtheta_{\mathrm{new}} = g_{\mathrm{feat}}(\bbtheta;\lambda)$ and labels $\bby_{\mathrm{new}} = g_{\mathrm{label}}(\bbtheta,\bby;\lambda)$ depend on the clusterpath in \eqref{e:graphmad_cp} for $\lambda\in[0,1]$.
We sample graphs $\{G_i^{\mathrm{new}}\}_{i=1}^{T'}$ from $\bbtheta_{\mathrm{new}}$ via \eqref{e:gmad_feat}.

To conclude our GraphMAD discussion, we note that while both data mixup \eqref{e:gmad_feat} and label mixup \eqref{e:gmad_label} depend on the clusterpath $\{\hat{\bbu}^{(k)}\}_{k=1}^K$ in \eqref{e:graphmad_cp}, these mixup functions need not be applied together. 
For example, for some $\lambda\in[0,1]$ and $k=1,\dots,K$, we may have $\bbtheta_{\mathrm{new}} = \hat{\bbu}^{(k)}(\lambda)$ from \eqref{e:gmad_feat} and $\bby_{\mathrm{new}} = \lambda\hat{\bby}^{(k)}(0) + (1-\lambda)\hat{\bby}^{(k)}(1)$ as in \eqref{e:orig_labelmix}.
In the next section we demonstrate GraphMAD data mixup via \eqref{e:gmad_feat} and label mixup via \eqref{e:gmad_label}, both implemented jointly with each other and with other choices of $g_{\mathrm{feat}}$ and $g_{\mathrm{label}}$. 

\begin{table}[t]
\scriptsize
\centering
\caption{\small{Counterpart of Table~\ref{t:molbio} for graph classification accuracy on social datasets.}}
\vspace{-2mm}
\begin{tabular}{ m{3.2em} m{5.2em} m{6em} m{6em} m{6em} }
\hline
\multicolumn{2}{c}{\bf Method} & \multicolumn{1}{c}{\bf COLLAB} & \multicolumn{1}{c}{\bf IMDB-B} & \multicolumn{1}{c}{\bf IMDB-M} \\
\hline
\multicolumn{2}{c}{Classes} & \multicolumn{1}{c}{3} & \multicolumn{1}{c}{2} & \multicolumn{1}{c}{3} \\
\multicolumn{2}{c}{Graphs} & \multicolumn{1}{c}{5000} & \multicolumn{1}{c}{1000} & \multicolumn{1}{c}{1500} \\
\hline
$g_{\mathrm{feat}}$ & $g_{\mathrm{label}}$ & & & \\
\hhline{|-|-|}
\textemdash & \textemdash & $\mathbf{80.00\pm0.96}$ & $ 73.14 \pm 3.15 $ & $ 47.71 \pm 4.25 $ \\[.07cm]
\multirow{4}{*}{Ln. \eqref{e:gmixup}} & Ln. \eqref{e:orig_labelmix}\cite{han2022g} & $77.60\pm1.53$ & $ 72.07 \pm 2.06 $ & $ 47.24 \pm 4.21 $ \\
& Sig. \eqref{e:mixup_label_sig} & $78.21\pm1.16$ & $ \mathbf{74.00 \pm 2.14} $ & $ \mathbf{49.67 \pm 2.15} $ \\
& Log. \eqref{e:mixup_label_log} & $78.19\pm1.61$ & $ 72.64 \pm 1.73 $ & $ 47.43 \pm 2.45 $ \\
& Cp. \eqref{e:gmad_label} & $78.41\pm0.99$ & $ 71.43 \pm 3.25 $ & $ 47.29 \pm 5.21 $ \\[.07cm]
\multirow{4}{*}{Cp. \eqref{e:gmad_feat}} & Ln. \eqref{e:orig_labelmix} & $78.93\pm2.63$ & $ 70.57 \pm 4.89 $ & $ 45.52 \pm 4.09 $ \\
& Sig. \eqref{e:mixup_label_sig} & $77.89\pm1.30$ & $ \mathbf{75.00 \pm 5.13} $ & $ 44.48 \pm 2.78 $ \\
& Log. \eqref{e:mixup_label_log} & $\mathbf{80.39\pm1.20}$ & $ 73.43 \pm 4.75 $ & $ 48.76 \pm 2.43 $ \\
& Cp. \eqref{e:gmad_label} & $79.55\pm2.29$ & $ 71.43 \pm 4.72 $ & $ \mathbf{49.71 \pm 4.33} $\\
\hline
\end{tabular}\label{t:social}
\vspace{-4mm}
\end{table}

\section{Results}\label{S:results}
\urlstyle{same}

We demonstrate GraphMAD for creating labeled graph data to improve graph classification performance. 
As a graph classifier, we use the same GNN for all our experiments, the Graph Isomorphism Network (GIN) \cite{xu2018powerful}, which uses node spatial relations to aggregate neighborhood features.
We compare classification performance for the original dataset with that of augmented datasets via multiple methods of graph mixup.

For mixup of graph data $g_{\mathrm{feat}}$, we compare GraphMAD's clusterpath data mixup \eqref{e:gmad_feat} with linear graphon mixup \cite{han2022g}.
In this latter approach, we estimate one SBM graphon $\bbW^{(k)}\in[0,1]^{D\times D}$ for each class $k=1,\dots, K$.
Then, for random pairs of classes $k,k'=1,\dots,K$ with $k\neq k'$, we linearly interpolate with $\lambda\in[0,1]$ and sample new graphs as
\begin{equation}\label{e:gmixup}
G_i^{\mathrm{new}} \sim \bbW_{\mathrm{new}} = \lambda\bbW^{(k)} + (1-\lambda)\bbW^{(k')}.
\end{equation}

We consider additional variants for the label mixup $g_{\mathrm{label}}$, apart from the classical linear function in~\eqref{e:orig_labelmix} and the proposed one in~\eqref{e:gmad_label}. In particular, given $\lambda\in[0,1]$ and label vectors $\bby_i,\bby_j\in[0,1]^K$, we also consider the mixup functions
\begin{subequations}\label{e:mixup_labels}
\begin{alignat}{3}
&
\bby_{\mathrm{new}} = \mathrm{sig}(\lambda)\bby_i + (1-\mathrm{sig}(\lambda))\bby_j, \label{e:mixup_label_sig}&\\
&
\bby_{\mathrm{new}} = \mathrm{logit}(\lambda)\bby_i + (1-\mathrm{logit}(\lambda))\bby_j \label{e:mixup_label_log},
&
\end{alignat}
\end{subequations}
where sigmoid refers to $\mathrm{sig}(x) = 1/\big(1+\exp\{-a(2x-1)\}\big)$ and logit is $\mathrm{logit}(x) = \mathrm{log}\big(x/(1-x)\big)/2a + 1/2$ for $a>0$ and $x\in[0,1]$.
Note that $\ccalG$-Mixup \cite{han2022g} has $g_{\mathrm{feat}}$ as \eqref{e:gmixup} and $g_{\mathrm{label}}$ as \eqref{e:orig_labelmix}

We perform graph classification on nine graph benchmark datasets from the TUDatasets collection \cite{morris2020tudataset}: DD, PROTEINS, and ENZYMES for protein categorization, AIDS, MUTAG, and NCI109 for molecule classification, and COLLAB, IMDB-B, and IMDB-M for social network classification. 
In Tables \ref{t:molbio} and \ref{t:social} we present classification accuracy for each dataset using all pairs of data and label mixup functions, where $g_{\mathrm{feat}}$ is \eqref{e:gmad_feat} or \eqref{e:gmixup}, and $g_{\mathrm{label}}$ is selected from \eqref{e:orig_labelmix}, \eqref{e:gmad_label}, or \eqref{e:mixup_labels}.
The first row with no choice of $g_{\mathrm{feat}}$ and $g_{\mathrm{label}}$ denotes performance using the original dataset.
The mixup parameter $\lambda\sim\text{Unif}([0,1])$ is the same for all choices of $g_{\mathrm{feat}}$ and $g_{\mathrm{label}}$.\footnote{Further implementation details can be found in our code, provided at \url{https://github.com/mn51/graphmad}.}

GraphMAD data mixup in \eqref{e:gmad_feat} achieves the best performance for all nine datasets, and \eqref{e:gmad_feat} is the only data augmentation method to improve performance for PROTEINS and COLLAB.
Furthermore, for all \emph{multi-class} datasets, since GraphMAD exploits relationships among all classes, it is always able to improve accuracy above the baseline compared to linear graphon mixup in \eqref{e:gmixup}, which only performs mixup between pairs of classes and thus is outperformed by the vanilla GIN for COLLAB and ENZYMES.

\begin{figure}
	\begin{minipage}[c]{.42\textwidth}
	    \centering
		\includegraphics[width=.8\textwidth]{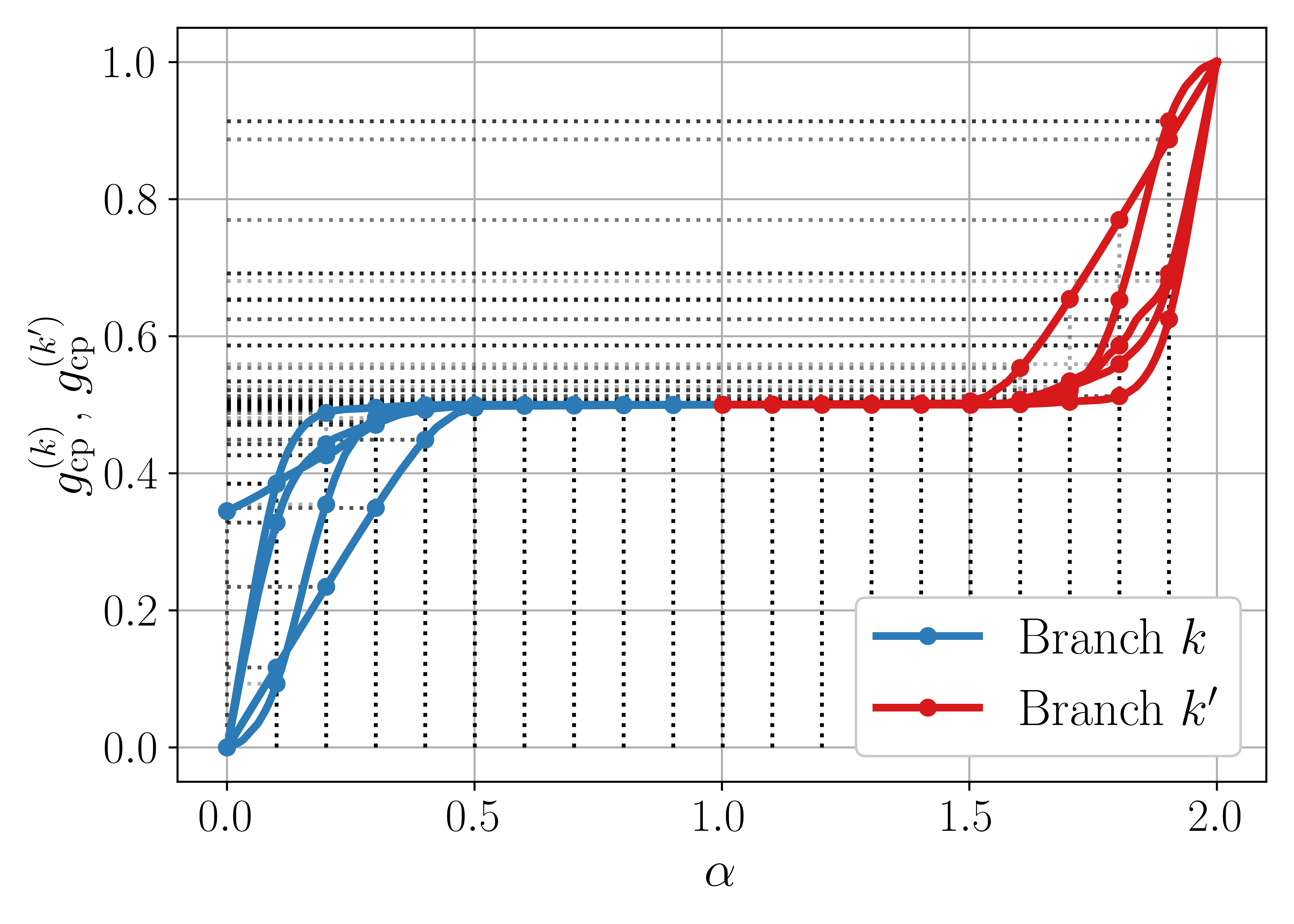}
	\end{minipage}
	\vspace{-2mm}
\caption{\small{Extended clusterpath behavior between branches $k$ and $k'$ for ENZYMES dataset. 
For $\alpha\in[0,2]$, we show the level of data mixup from branch $k$ to fusion in blue as $\frac{1}{2}{g}_{\mathrm{cp}}^{(k)}(\alpha)$ for $\alpha\in[0,1]$ and from fusion to branch $k'$ in red as $1-\frac{1}{2}g_{\mathrm{cp}}^{(k')}(2-\alpha)$ for $\alpha\in[1,2]$.
}}
	\vspace{-4mm}
\label{fig:enzymes_cp}
\end{figure}

We examine the 6-class ENZYMES dataset, whose performance using GraphMAD data mixup in \eqref{e:gmad_feat} and sigmoid label mixup in \eqref{e:mixup_label_sig} achieves an accuracy gap of around $6\%$ over other methods.
For topology-sensitive protein datasets such as ENZYMES, new data close in graphon space to the original data may produce incorrect graph structures for protein categorization, hence we wish to sample farther away from the original classes.
This is automatically captured by the shape of the clusterpath.
The data clusterpath across two branches $\hat{\bbu}^{(k)}$ and $\hat{\bbu}^{(k')}$ is shown in Fig.~\ref{fig:enzymes_cp} for several choices of $k$ and $k'$.
For uniformly selected values of $\lambda$, new data $\bbtheta_{\mathrm{new}}$ are close to the mean of the dataset for several values of $\lambda$.
Thus samples generated through the data-driven GraphMAD are generated far from the original data, which dramatically improves performance.

\section{Conclusion}\label{S:concl}

We proposed a data-driven mixup function for graph data augmentation via convex clustering.
Potential directions for development include learning parameters for convex clustering, such as the fusion weights and the selected clusterpath branch for sampling.
Furthermore, we can convert GraphMAD to a completely data-driven approach by using similarity-based fusion weights and applying self-supervised learning approaches.

		
		

\bibliographystyle{ieeetr}
\bibliography{citations}

\begin{thebibliography}{10}

\bibitem{gilmer2017neural}
J.~Gilmer, S.~S. Schoenholz, P.~F. Riley, O.~Vinyals, and G.~E. Dahl, ``Neural
  message passing for quantum chemistry,'' in {\em Intl. Conf. on Machine
  Learning (ICML)}, pp.~1263--1272, PMLR, 2017.

\bibitem{chowdhury2021unfolding}
A.~Chowdhury, G.~Verma, C.~Rao, A.~Swami, and S.~Segarra, ``Unfolding {WMMSE}
  using graph neural networks for efficient power allocation,'' {\em IEEE
  Trans. Wireless Commun.}, vol.~20, no.~9, pp.~6004--6017, 2021.

\bibitem{qiu2018deepinf}
J.~Qiu, J.~Tang, H.~Ma, Y.~Dong, K.~Wang, and J.~Tang, ``{DeepInf}: Social
  influence prediction with deep learning,'' in {\em Intl. Conf. on Knowledge
  Discovery and Data Mining (SIGKDD)}, pp.~2110--2119, 2018.

\bibitem{lewy2022overview}
D.~Lewy and J.~Ma{\'n}dziuk, ``An overview of mixing augmentation methods and
  augmentation strategies,'' {\em Artif. Intell. Review}, pp.~1--59, 2022.

\bibitem{guo2019augmenting}
H.~Guo, Y.~Mao, and R.~Zhang, ``Augmenting data with mixup for sentence
  classification: An empirical study,'' {\em arXiv preprint arXiv:1905.08941},
  2019.

\bibitem{zhang2017mixup}
H.~Zhang, M.~Cisse, Y.~N. Dauphin, and D.~Lopez-Paz, ``{mixup}: Beyond
  empirical risk minimization,'' {\em arXiv preprint arXiv:1710.09412}, 2017.

\bibitem{guo2019mixup}
H.~Guo, Y.~Mao, and R.~Zhang, ``Mixup as locally linear out-of-manifold
  regularization,'' in {\em AAAI Conf. on Artif. Intell.}, vol.~33,
  pp.~3714--3722, 2019.

\bibitem{zhang2021does}
L.~Zhang, Z.~Deng, K.~Kawaguchi, A.~Ghorbani, and J.~Zou, ``How does mixup help
  with robustness and generalization?,'' in {\em Intl. Conf. on Learning
  Representations (ICLR)}, 2021.

\bibitem{guo2021intrusion}
H.~Guo and Y.~Mao, ``{ifMixup}: Towards intrusion-free graph mixup for graph
  classification,'' {\em arXiv preprint arXiv:2110.09344}, 2021.

\bibitem{park2021graph}
J.~Park, H.~Shim, and E.~Yang, ``{Graph Transplant}: Node saliency-guided graph
  mixup with local structure preservation,'' {\em arXiv preprint
  arXiv:2111.05639}, 2021.

\bibitem{ding2022data}
K.~Ding, Z.~Xu, H.~Tong, and H.~Liu, ``Data augmentation for deep graph
  learning: A survey,'' {\em arXiv preprint arXiv:2202.08235}, 2022.

\bibitem{zhao2022graph}
T.~Zhao, G.~Liu, S.~G{\"u}nnemann, and M.~Jiang, ``Graph data augmentation for
  graph machine learning: A survey,'' {\em arXiv preprint arXiv:2202.08871},
  2022.

\bibitem{wang2021mixup}
Y.~Wang, W.~Wang, Y.~Liang, Y.~Cai, and B.~Hooi, ``Mixup for node and graph
  classification,'' in {\em Proc. World Wide Web Conf.}, pp.~3663--3674, 2021.

\bibitem{verma2021graphmix}
V.~Verma, M.~Qu, K.~Kawaguchi, A.~Lamb, Y.~Bengio, J.~Kannala, and J.~Tang,
  ``{GraphMix}: Improved training of {GNNs} for semi-supervised learning,'' in
  {\em AAAI Conf. on Artif. Intell.}, vol.~35, pp.~10024--10032, 2021.

\bibitem{han2022g}
X.~Han, Z.~Jiang, N.~Liu, and X.~Hu, ``{G-Mixup}: Graph data augmentation for
  graph classification,'' {\em arXiv preprint arXiv:2202.07179}, 2022.

\bibitem{guo2020nonlinear}
H.~Guo, ``Nonlinear mixup: Out-of-manifold data augmentation for text
  classification,'' in {\em AAAI Conf. on Artif. Intell.}, vol.~34,
  pp.~4044--4051, 2020.

\bibitem{mai2021metamixup}
Z.~Mai, G.~Hu, D.~Chen, F.~Shen, and H.~T. Shen, ``{MetaMixUp}: Learning
  adaptive interpolation policy of mixup with metalearning,'' {\em IEEE Trans.
  Neural Netw. and Learning Sys.}, 2021.

\bibitem{lovasz2012large}
L.~Lov{\'a}sz, {\em Large networks and graph limits}, vol.~60.
\newblock American Mathematical Soc., 2012.

\bibitem{diaconis2007graph}
P.~Diaconis and S.~Janson, ``Graph limits and exchangeable random graphs,''
  {\em arXiv preprint arXiv:0712.2749}, 2007.

\bibitem{avella2020centrality}
M.~{Avella-Medina}, F.~{Parise}, M.~T. {Schaub}, and S.~{Segarra}, ``Centrality
  measures for graphons: Accounting for uncertainty in networks,'' {\em IEEE
  Trans. Network Science and Eng.}, vol.~7, no.~1, pp.~520--537, 2020.

\bibitem{sischka2022based}
B.~Sischka and G.~Kauermann, ``{EM}-based smooth graphon estimation using
  {MCMC} and spline-based approaches,'' {\em Social Networks}, vol.~68,
  pp.~279--295, 2022.

\bibitem{chan2014consistent}
S.~Chan and E.~Airoldi, ``A consistent histogram estimator for exchangeable
  graph models,'' in {\em Intl. Conf. on Machine Learning (ICML)},
  pp.~208--216, PMLR, 2014.

\bibitem{chan2013estimation}
S.~H. Chan, T.~B. Costa, and E.~M. Airoldi, ``Estimation of exchangeable graph
  models by stochastic blockmodel approximation,'' in {\em IEEE Global Conf.
  Signal and Info. Process. (GlobalSIP)}, pp.~293--296, IEEE, 2013.

\bibitem{pelckmans2005convex}
K.~Pelckmans, J.~De~Brabanter, J.~A. Suykens, and B.~De~Moor, ``Convex
  clustering shrinkage,'' in {\em Statistics and Optimization of Clustering
  Workshop (PASCAL)}, 2005.

\bibitem{hocking2011clusterpath}
T.~D. Hocking, A.~Joulin, F.~Bach, and J.-P. Vert, ``Clusterpath: An algorithm
  for clustering using convex fusion penalties,'' in {\em Intl. Conf. on
  Machine Learning (ICML)}, p.~1, 2011.

\bibitem{lindsten2011clustering}
F.~Lindsten, H.~Ohlsson, and L.~Ljung, ``Clustering using sum-of-norms
  regularization: With application to particle filter output computation,'' in
  {\em IEEE Wrkshp. Statistical Signal Process. (SSP)}, pp.~201--204, IEEE,
  2011.

\bibitem{ohlsson2010segmentation}
H.~Ohlsson, L.~Ljung, and S.~Boyd, ``Segmentation of {ARX}-models using
  sum-of-norms regularization,'' {\em Automatica}, vol.~46, no.~6,
  pp.~1107--1111, 2010.

\bibitem{tan2015statistical}
K.~M. Tan and D.~Witten, ``Statistical properties of convex clustering,'' {\em
  Electronic Journal of Statistics}, vol.~9, no.~2, p.~2324, 2015.

\bibitem{zhu2014convex}
C.~Zhu, H.~Xu, C.~Leng, and S.~Yan, ``Convex optimization procedure for
  clustering: Theoretical revisit,'' {\em Advances in Neural Information
  Processing Systems}, vol.~27, 2014.

\bibitem{chi2019recovering}
E.~C. Chi and S.~Steinerberger, ``Recovering trees with convex clustering,''
  {\em SIAM Journal on Mathematics of Data Science}, vol.~1, no.~3,
  pp.~383--407, 2019.

\bibitem{choi2019convex}
H.~Choi and S.~Lee, ``Convex clustering for binary data,'' {\em Advances in
  Data Analysis and Classification}, vol.~13, no.~4, pp.~991--1018, 2019.

\bibitem{navarro2021network}
M.~Navarro, G.~I. Allen, and M.~Weylandt, ``Network clustering for latent state
  and changepoint detection,'' {\em arXiv preprint arXiv:2111.01273}, 2021.

\bibitem{kipf2016semi}
T.~N. Kipf and M.~Welling, ``Semi-supervised classification with graph
  convolutional networks,'' {\em arXiv preprint arXiv:1609.02907}, 2016.

\bibitem{xu2018powerful}
K.~Xu, W.~Hu, J.~Leskovec, and S.~Jegelka, ``How powerful are graph neural
  networks?,'' {\em arXiv preprint arXiv:1810.00826}, 2018.

\bibitem{morris2020tudataset}
C.~Morris, N.~M. Kriege, F.~Bause, K.~Kersting, P.~Mutzel, and M.~Neumann,
  ``{TUDataset}: A collection of benchmark datasets for learning with graphs,''
  {\em arXiv preprint arXiv:2007.08663}, 2020.

\end{thebibliography}

\end{document}